\documentclass{article}
\usepackage{spconf,amsmath,graphicx} 

\usepackage{amssymb,bm}

\usepackage{multirow,bigstrut,tabularx,threeparttable}

\usepackage{hyperref}

\usepackage{subfigure,subcaption}

\usepackage{algorithm,algorithmic}

\usepackage[top=2cm, bottom=2cm, left=2cm, right=2cm]{geometry}

\usepackage{balance}

\usepackage{booktabs}

\title{Synthesizing Black-box Anti-forensics DeepFakes with High Visual Quality}
\name{Bing Fan$^1$ \qquad Shu Hu$^2$ \qquad Feng Ding$^1$$^{\star}$\thanks{$^{\star}$Corresponding author}
\thanks{This work was supported in part by the National Natural Science Foundation of China under Grant 62262041, and in part by the Jiangxi Provincial Natural Science Foundation under Grant 20232BAB202011.}
\thanks{This paper is accepted by ICASSP 2024.}}
\address{$^1$Nanchang University \qquad $^2$Purdue University in Indianapolis}

\begin{document}
\maketitle
\begin{abstract}
DeepFake, an AI technology for creating facial forgeries, has garnered global attention. Amid such circumstances, forensics researchers focus on developing defensive algorithms to counter these threats. In contrast, there are techniques developed for enhancing the aggressiveness of DeepFake, e.g., through anti-forensics attacks, to disrupt forensic detectors. However, such attacks often sacrifice image visual quality for improved undetectability. To address this issue, we propose a method to generate novel adversarial sharpening masks for launching black-box anti-forensics attacks. Unlike many existing arts, with such perturbations injected, DeepFakes could achieve high anti-forensics performance while exhibiting pleasant sharpening visual effects. After experimental evaluations, we prove that the proposed method could successfully disrupt the state-of-the-art DeepFake detectors. Besides, compared with the images processed by existing DeepFake anti-forensics methods, the visual qualities of anti-forensics DeepFakes rendered by the proposed method are significantly refined.
\end{abstract}
\begin{keywords}
Multimedia forensics, DeepFake, anti-forensics, image processing, adversarial sharpening mask.
\end{keywords}
\section{Introduction}
\label{sec:intro}

In the era of digital media, images and videos play a crucial role. However, the rapid advancements in AI technology have given rise to a concerning phenomenon known as DeepFakes, which are highly convincing digital forgeries\cite{lyu2022deepfake}. These DeepFakes pose significant security and privacy threats. Numerous researchers have made significant contributions \cite{li2022artifacts, ding2022securing, zhang2023x, yang2023improving, fan2023attacking, ju2023improving, pu2022learning, guo2022open, guo2022eyes, guo2022robust, wang2022gan, hu2021exposing} to addressing the challenges posed by DeepFake technology and uncovering facial forgeries. With the advancement of deep learning, these forensic methods often rely on deep neural networks (CNN) to frame the problem as binary classification, distinguishing between real and DeepFake images\cite{zhou2023exposing}. Many of these methods have shown remarkable forensic capabilities on benchmark datasets, employing popular classification networks like ResNet\cite{he2016deep}, EfficientNet\cite{tan2019efficientnet}, DenseNet\cite{huang2017densely}, and ConvNet\cite{liu2022convnet}. Additionally, there have been efforts to improve the robustness of deepfake detectors. For instance, SBIs \cite{shiohara2022detecting} utilizes a self-blended algorithm to learn from synthetic image processing, reducing the reliance on specific DeepFake datasets and compensating for the lack of data subsets in fake data processing.

At the same time, significant progress has been made in DeepFake anti-forensics aiming to hide DeepFake images by injecting adversarial noise, making it challenging for forensic detectors to detect them. However, current anti-forensics methods often lead to severe visual quality degradation, especially in facial images, where humans can easily spot noticeable artifacts and inconsistencies \cite{wang2021adversarial}. Successful anti-forensics samples must pass scrutiny from both forensic algorithms and human observers \cite{ding2022exs}. To address this challenge, the key is to maintain acceptable visual quality while introducing adversarial perturbations. GAN models have been employed for this purpose, focusing on reducing noticeable visual artifacts caused by injecting perturbations, but this can introduce detectable traces and reduce undetectability, posing a challenge in balancing undetectability and image visual quality in DeepFake anti-forensics \cite{ding2021anti}.



\begin{figure}[!t]
\centering
\includegraphics [width=8.5cm]{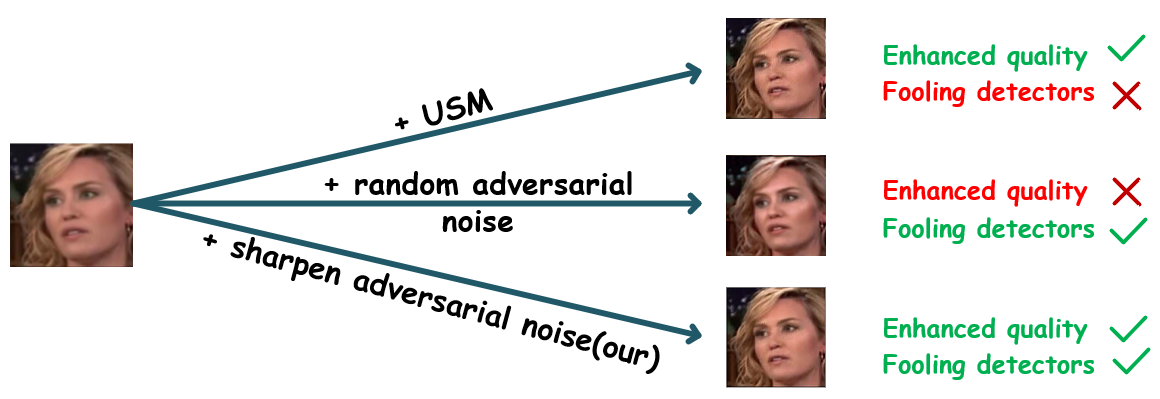}
\caption{illustrate our motivation}\label{illus_motivation}
\end{figure}

Hence, in this paper, we present a novel DeepFake anti-forensics framework that addresses the challenges mentioned above. Unlike existing methods, our approach does not focus on reducing the perceptible effects of adversarial noises. Instead, we introduce more perturbations but constrain them to make DeepFake images appear as if they were processed with a sharpening mask, which is illustrated in Fig. \ref{illus_motivation}. Our main contributions include the following: 
\begin{enumerate}
  \item Introduction of a method to generate adversarial sharpening masks, resulting in DeepFake images that are highly undetectable and visually appealing. 
  \item Adoption of a parameter-frozen training strategy and the integration of MobileVit blocks\cite{mehta2021mobilevit} into the VEN to produce higher-quality anti-forensics images.
  \item Evaluation of the propose approach through comparisons with state-of-the-art DeepFake anti-forensics methods on various benchmark datasets.
\end{enumerate}




\section{method}
\label{sec:format}
\begin{figure*}[!t]
\centering
\includegraphics [width=14cm]{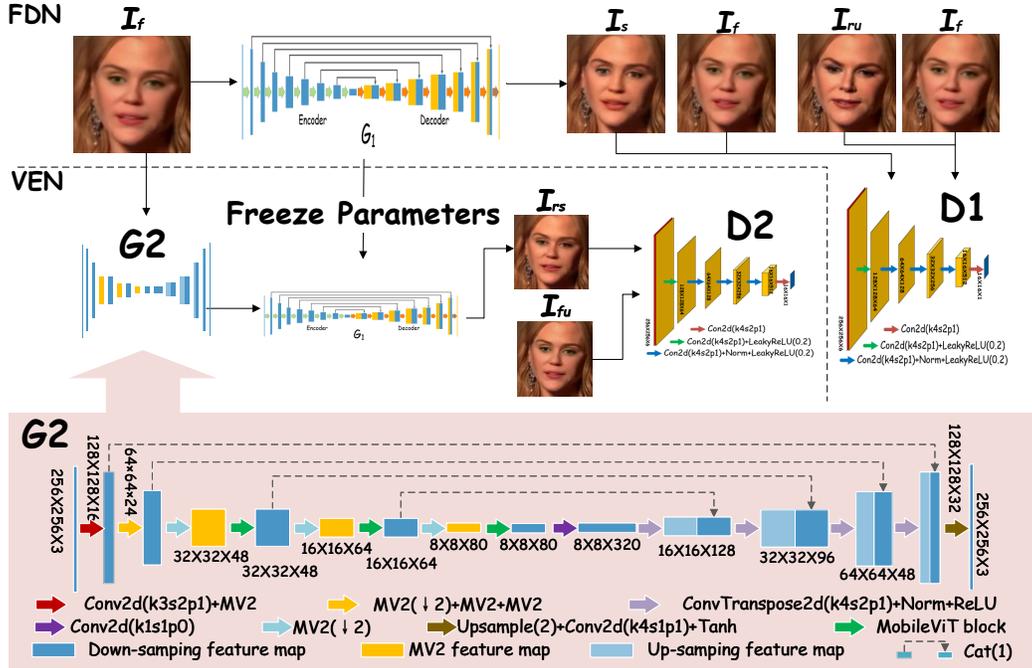}
\caption{The framework of our proposed method.}\label{framework}
\end{figure*}

\textbf{Preliminaries.} We proposed a method that consists of two separate modules, shown in Fig. \ref{framework}. 

\textit{The first module is the Forensics Disruption Network (FDN), where our primary goal is to ensure that the synthesized anti-forensics images can mislead DeepFake detectors through the injection of adversarial noise and achieve high undetectability. }At this stage, we can tolerate lower visual quality in the images. After training FDN, the FDN network can generate highly undetectable synthesizing images denoted as $I_s = I_f + m$, where $I_f$ is the input DeepFake image, and $m$ is the adversarial mask. 

\begin{figure}[!t]
\centering
\includegraphics [width=8.5cm]{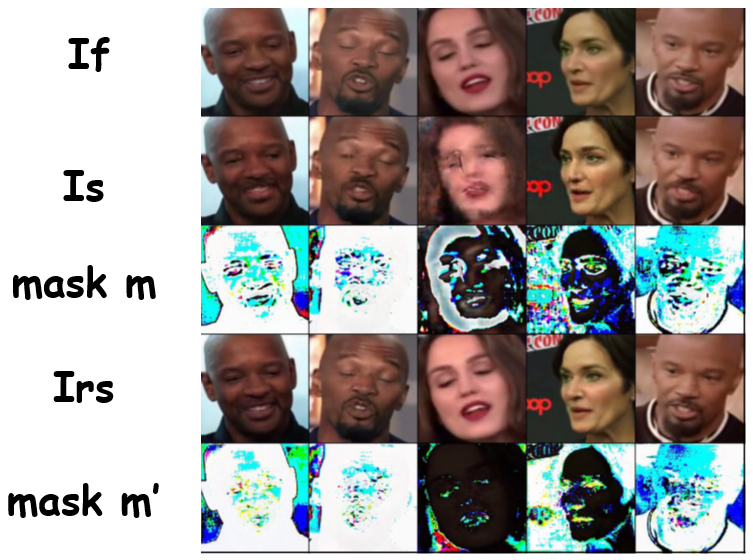}
\caption{Samples, first row: original DeepFake frames; second row: images synthesized by FDN; third row: adversarial mask $m$; fourth row: re-synthesized by VEN; fifth row:  sharpening adversarial mask $m'$.}\label{sample_FDN_VEN}
\end{figure}

Furthermore, in the second stage, \textit{we introduce the Visual Enhancement Network (VEN), which includes MobileVit blocks\cite{mehta2021mobilevit}, to fine-tune and compensate for the visual quality loss caused by the injection of adversarial perturbations in the FDN network.}
Specifically, we initialize the parameters of G1 in the VEN network with the parameters of G1 trained in the first step of FDN and freeze it for subsequent training. Then, we fine-tune the parameters of G2 by training the VEN network. Finally, the trained VEN network is capable of re-synthesizing images with a sharpening effect and high undetectability, denoted as $I_rs = I_f + m'$, where m' represents the sharpening adversarial mask. A visual representation of some sample results can be seen in Fig. \ref{sample_FDN_VEN} for a more intuitive understanding.

\vspace{-5mm}

\subsection{Forensics Disruption Network}

To pursue high adversarial performance for $I_s$, we define the loss function of the generator $G_1$ as $\mathcal{L}_{G_1}$. It comprises the generating loss $\mathcal{L}_{gan1}$ and the reconstruction loss $\mathcal{L}_{re1}$,
\begin{equation}
\begin{aligned}
\mathcal L_{gan 1} &=   E\left[1-log D_1\left(I_f \oplus I_s\right)\right] \\ &=  E\left[1-log D_1\left(I_f \oplus G_1(I_f)\right)\right],\\
\mathcal L_{re1} &={||I_s-I_{ru}||}_1={||G_1(I_f)-I_{ru}||}_1,
\end{aligned}
\end{equation}
where $I_{ru}$ represents the real images sharpened using the Unsharp Mask (USM) technique, and the symbol $\oplus$  refers to channel-wise concatenation. The purpose of designing $L_gan1$ is to learn features from real images to enhance undetectability. While ensuring high undetectability, we use an L1 norm loss to calculate $L_{re1}$ with the aim of minimizing the pixel-wise differences between $I_{ru}$ and $I_s$. In summary, $\mathcal{L}_{G_1}$ is 

\begin{equation}
\begin{aligned}
\min_{\theta G_1}{\mathcal L_{G_1}}=\mathcal L_{gan1}+\alpha \mathcal L_{re1},
\end{aligned}
\end{equation}
where $\alpha$ is a weight factor, and $\theta G_1$ denotes the parameters of $G_1$. 

On the other hand, the loss function for $D_1$ is defined as:

\begin{equation}
\begin{aligned}
\max_{\theta D_1}{\mathcal L_{D_1}} =& E\left[1-log D_1\left(I_f \oplus G\left(I_f\right)\right)\right] \\ & +E\left[log D_1\left(I_f \oplus I_{ru}\right)\right],
\end{aligned}
\end{equation}
where $\theta D_1$ denotes the parameters of $D_1$.
The total loss in FDN can be organized as:

\begin{equation}
\begin{aligned}
T_1= & \arg{\max_{\theta D_1}{\min_{\theta G_1} \mathcal L_{FND}}} \\ = & \arg{\max_{\theta D_1}{\min_{\theta G_1} \left( \mathcal  L_{G_1} + \mathcal L_{D_1} \right) }}.
\end{aligned}
\end{equation}

\subsection{Visual Enhancement Network}

Similar to FDN, we use the generating loss $\mathcal{L}{gan2}$ and reconstruction loss $\mathcal{L}{re2}$ as components of the overall loss function for $G_2$. This ensures that the synthesized image $I_{rs}$ in VEN has a distribution similar to that of $I_{fu}$, which represents Deepfake images sharpened using USM. This similarity benefits the restoration of details in $I_{rs}$. We define $\mathcal{L}_{gan2}$ as:
\begin{equation}
\begin{aligned}
\mathcal L_{gan2}= & E\left[1-log D_2\left(I_{rs}\right)\right] \\ = & E\left[1-log D_2\left(G_1\left(G_2\left(I_f\right)\right)\right)\right].
\end{aligned}
\end{equation}

The reconstruction loss is also the L1-norm to optimize the adversarial sharpening mask $m'$ for $I_{rs}$ by refining the sharpening effect. It can be defined as 
\begin{equation}
\begin{aligned}
\mathcal L_{re2}={||I_{rs}-I_{fu}||}_1={||{G_1(G}_2\left(I_f\right))-I_{fu}||}_1.
\end{aligned}
\end{equation}

Therefore, the general loss function for $G_2$ is
\begin{equation}
\begin{aligned}
\min_{\theta G_2}{\mathcal L_{G_2}}=\mathcal L_{gan2}+\beta \mathcal L_{re2},
\end{aligned}
\end{equation}
where $\beta$ is a weighting factors, and $\theta G_2$ is the parameters of $G_2$. Note that the $G_1$ is pre-trained in FDN. The parameters learned in FDN are frozen for training VEN. Hence, the complete form for $G_2$ can be expressed by

\begin{equation}
\begin{aligned}
\min_{\theta G_1=p, \theta G_2} {\mathcal L_{G_2}} = & E\left[1-log D_2\left(G_1\left(G_2\left(I_f\right)\right)\right)\right] \\ + & \beta{||{G_1(G}_2\left(I_f\right))-I_{fu}||}_1,
\end{aligned}
\end{equation}
where $p$ denotes the parameters learned in FDN, and $\theta G_2$ is the parameter we intend to derive by minimizing $\mathcal{L}_{G_2}$ during training VEN. The loss function for $D_2$ can be defined as:
\begin{equation} 
\begin{aligned} 
\max _ { \theta D_2 }{ \mathcal  L_{D_2}}  =  & E \left[  1-log D_2 \left( G_1 { (G}_2 \left( I_f \right) ) \right) \right] \\ 
+ & E\left[log D_2\left(I_{fu}\right)\right].
\end{aligned}
\end{equation}
In summary, the target for training the entire VEN can be organized as:

\begin{equation}
\begin{aligned}
T_2= & \arg{{\max_{\theta D_2}{\min_{\theta G_2}{\mathcal{L}_{VEN}}}}} \\ = &  \arg{{\max_{\theta D_2}{\min_{\theta G_2}{ \left( \mathcal{L}_{G_2} + \mathcal{L}_{D_2} \right) } }}}.
\end{aligned}
\end{equation}

In addition, we attempt to utilize a single GAN architecture incorporating both the forensics loss and the visual loss. However, due to the inherent conflict between these losses, finding a balance between forensics performance and visual quality proves challenging.
Notably, we also investigate the order of G1 and G2 within VEN. Our experiments reveal that when G1 precedes G2, the resulting images exhibit superior visual quality but limited effectiveness in disrupting forensics detectors. To address this issue, we utilize G2 as the pre-processor for G1 and fine-tune G2 during training to achieve the balance.

\begin{table}[!ht]  
  \centering  
  \fontsize{6}{7}\selectfont  
  \begin{threeparttable}  
  \caption{Classification Accuracy and Prediction Precision for identity detectors} 
  \label{table_detector_accuracy}  
    \begin{tabular}{@{}c@{\hspace{4pt}}c@{\hspace{4pt}}c@{\hspace{4pt}}c@{\hspace{4pt}}c@{\hspace{4pt}}c@{\hspace{4pt}}c@{}}  
    \toprule  
    \multirow{2}{*}{Detectors}&  
    \multicolumn{3}{c}{Accuracy}&\multicolumn{3}{c}{Prediction}\cr  
    \cmidrule(lr){2-4} \cmidrule(lr){5-7}  
     & Celeb-DF & DeeperForensics & FF++ & Celeb-DF & DeeperForensics & FF++
     \cr
    \midrule 
    ResNet-50 \cite{he2016deep}   & 98.66\% & 99.48\% & 98.99\% & 99.08\% & 99.59\% & 99.33\% \\
    DenseNet-121 \cite{huang2017densely} & 98.79\% & 99.67\% & 99.39\% & 99.21\% & 99.64\% & 99.29\% \\
    EfficientNet \cite{tan2019efficientnet}  & 98.65\% & 99.72\% & 99.21\% & 98.89\% & 99.76\% & 99.16\% \\
    MobileNet   \cite{howard2017mobilenets}  & 97.12\% & 99.24\% & 98.56\% & 97.52\% & 99.20\% & 98.32\% \\
    ShuffleNet \cite{zhang2018shufflenet}   & 98.45\% & 99.68\% & 99.17\% & 98.75\% & 99.75\% & 99.18\% \\
    ConvNeXt   \cite{liu2022convnet}  & 98.64\% & 99.53\% & 99.11\% & 98.78\% & 99.46\% & 99.51\% \\
    EfficientNet-SBIs \cite{shiohara2022detecting} & 93.18\% & - & 99.64\% & 92.65\% & - & 99.32\% \\
    \bottomrule  
    \end{tabular}  
    \end{threeparttable}  
\end{table}

\begin{figure}[!t]
	\centering
	\subfigure{\includegraphics[width=.49\columnwidth]{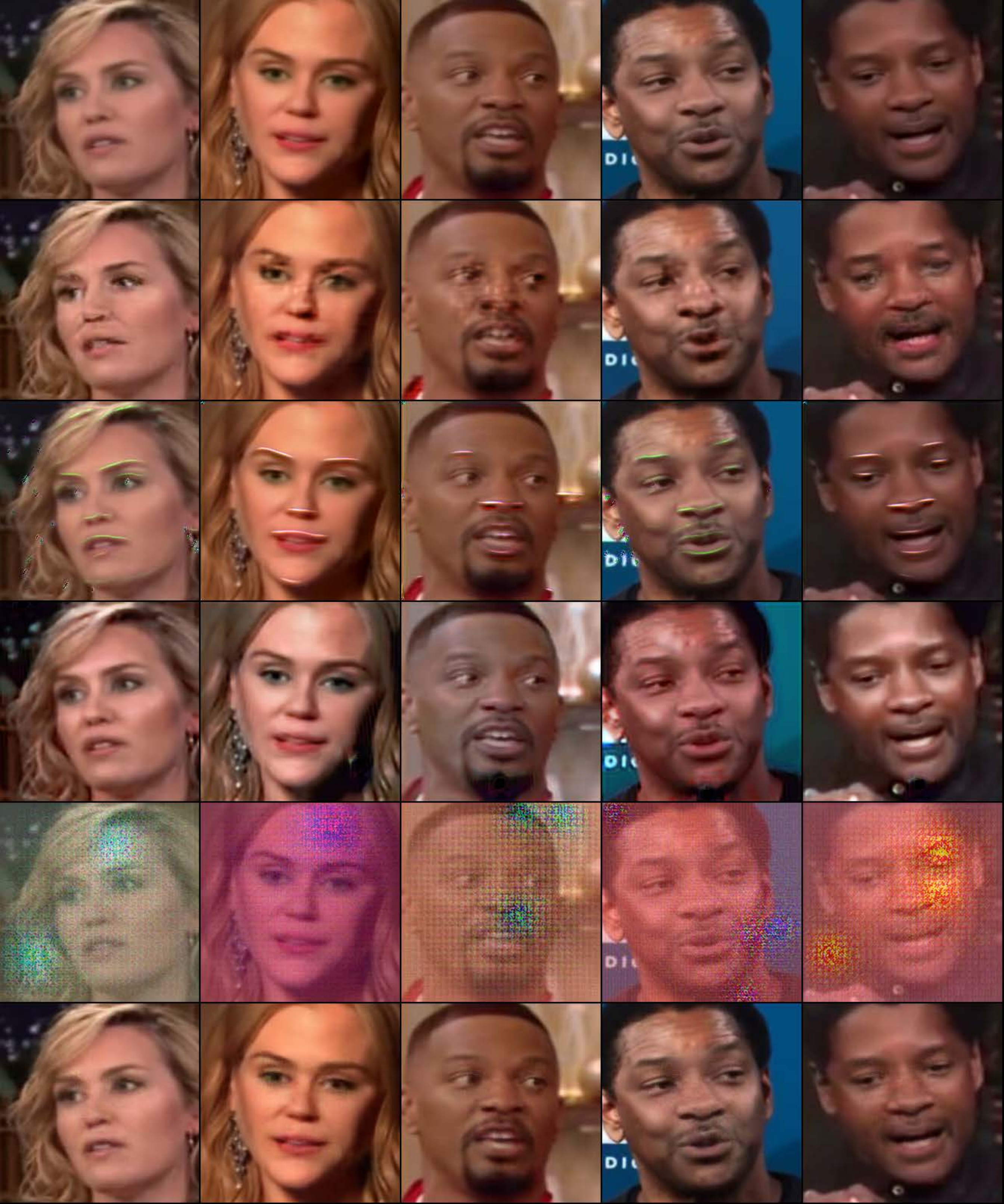}}\hspace{1pt}
	\subfigure{\includegraphics[width=.49\columnwidth]{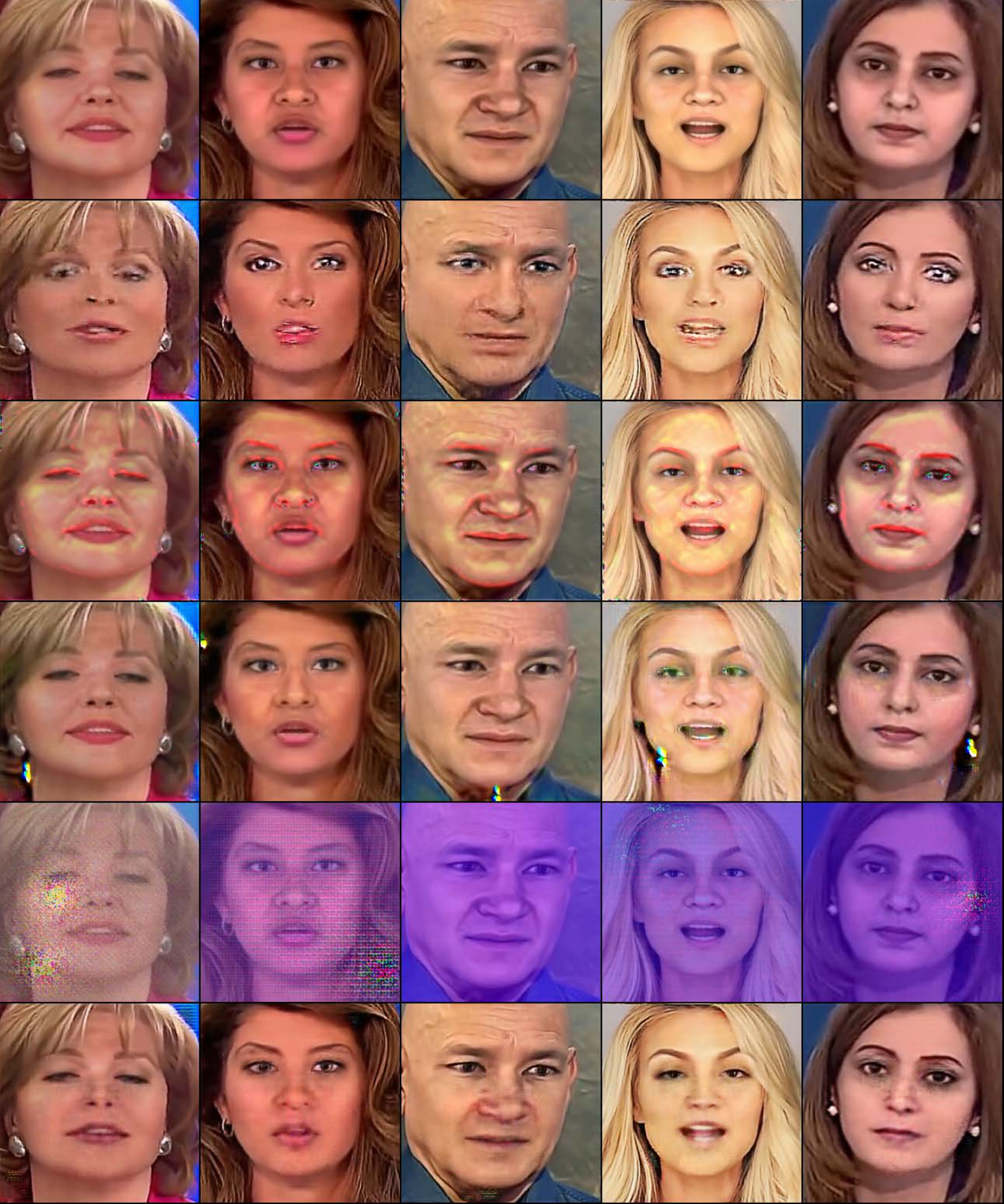}}
	\caption{Samples from Celeb-DF(left) and DeeperForensics-1.0(right). First row: original DeepFakes. Second row: \cite{isola2017image}. Third row:\cite{wang2021adversarial}. Fourth row:  \cite{ding2021anti}. Fifth row: \cite{fan2021deepfake}. Sixth row: proposed method. }\label{compare_Visual_img}
\end{figure}

\begin{table*}[!t]  
  \centering  
  \fontsize{6.5}{7.5}\selectfont  
  \begin{threeparttable}  
  \captionsetup{font={large}}
  \caption{Prediction precision for images synthesized by different methods} 
  \label{compare_undetected}  
    \begin{tabular}{@{}c@{\hspace{2pt}}c@{\hspace{2pt}}c@{\hspace{2pt}}c@{\hspace{2pt}}c@{\hspace{2pt}}c@{\hspace{2pt}}c@{\hspace{2pt}}c@{\hspace{2pt}}c@{\hspace{2pt}}c@{\hspace{2pt}}c@{\hspace{2pt}}c@{\hspace{2pt}}c@{\hspace{2pt}}c@{\hspace{2pt}}c@{\hspace{2pt}}c@{\hspace{2pt}}c@{\hspace{2pt}}c@{\hspace{2pt}}c@{\hspace{2pt}}c@{\hspace{2pt}}c@{\hspace{2pt}}c@{}}  
    \toprule  
    \multirow{2}{*}{Model}&  
    \multicolumn{7}{c}{Celeb-DF}&\multicolumn{7}{c}{DeeperForensics}&\multicolumn{7}{c}{FF++}\cr  
    \cmidrule(lr){2-8} \cmidrule(lr){9-15} \cmidrule(lr){16-22} 
     &\cite{isola2017image}$\downarrow$ &\cite{ding2021anti}$\downarrow$ &\cite{fan2021deepfake}$\downarrow$ &\cite{wang2021adversarial}$\downarrow$ &ours $\downarrow$ &$I_f$ $\uparrow$ &$I_{fu}$ $\uparrow$
     &\cite{isola2017image}$\downarrow$ &\cite{ding2021anti}$\downarrow$ &\cite{fan2021deepfake}$\downarrow$ &\cite{wang2021adversarial}$\downarrow$ &ours $\downarrow$ &$I_f$ $\uparrow$ &$I_{fu}$ $\uparrow$
     &\cite{isola2017image}$\downarrow$ &\cite{ding2021anti}$\downarrow$ &\cite{fan2021deepfake}$\downarrow$ &\cite{wang2021adversarial}$\downarrow$ &ours $\downarrow$ &$I_f$ $\uparrow$ &$I_{fu}$ $\uparrow$
     \cr
    \midrule 
    ResNet             & 6.28 \%& 8.83\% & 85.57\% & 19.49\% & \textbf{4.26\%} & 99.08\% & 98.27\%   &  7.67\%  & 23.88\%          & 24.21\% & 27.49\% & \textbf{4.71\%} & 99.59\%   & 97.35\%    &  9.42\% & 12.72\%         & 41.23\%  & 10.65\%         & \textbf{4.98\%} & 99.33\%   & 99.13\%  \cr
DenseNet           & 3.97 \%& 11.79\% & 15.16\% & \textbf{3.25\%} & 3.43 \% & 99.21\% & 96.44\%   & 15.24\%  & 24.15\%          & 22.83\% & 25.35\% & \textbf{8.54\%} & 99.64\%   & 99.62\%    & \textbf{1.78\%} & 11.83\%         & 66.43\%  & 4.03\%   & 6.43 \%  & 99.29\%   & 99.59\%  \cr
EfficientNet       & 7.11 \%& 17.70\% & 20.32\% & 29.32\% & \textbf{6.36\%} & 98.89\% & 98.48\%   & 10.05\%  & 37.46\%          & 18.07\% & 29.87\% & \textbf{4.65\%} & 99.76\%   & 99.17\%    & \textbf{4.71\%}& 9.15\%   & 33.47\%  & 26.82\%         & 41.01 \%  & 99.16\%   & 97.57\%  \cr
MobileNet          & 8.64 \%& 29.30\% & 11.45\% & 10.27\% & \textbf{3.18\%} & 97.52\% & 97.74\%   & 35.21\%  & \textbf{31.16\%} & 46.60\% & 36.57\% & 36.80 \%  & 99.20\%   & 97.36\%    & 27.19\%        & 21.98\%         & 51.88\%  & 15.68\%   & \textbf{1.77\%} & 98.32\%   & 95.28\%  \cr
ShuffleNet         & 13.67\%& 26.53\% & 19.31\% & 17.84\% & \textbf{2.64\%} & 98.75\% & 97.53\%   &  9.17\% & 15.88\%          & 17.73\% & 20.45\% & \textbf{2.25\%} & 99.75\%   & 98.74\%    & \textbf{4.21\%}& 22.52\%         & 26.05\%  & 6.16\%     & 15.50 \%  & 99.18\%   & 95.37\%  \cr
ConvNeXt           & 11.50\%& 30.28\% & 13.45\% & 17.60\% & \textbf{4.42\%} & 98.78\% & 99.28\%   & \textbf{5.37\%}  & 30.61\%          & 32.84\% & 30.63\% & 8.21\% & 99.46\%   & 99.56\%    & 14.18\%  & 29.68\%         & 25.36\%  & 19.62\%         & \textbf{4.91\%} & 99.51\%   & 98.67\%  \cr
EfficientNet-SBIs  & 24.53\%& 41.49\% & 26.89\% & 25.24\% & \textbf{12.33\%} & 92.65\% & 91.48\%    & -  & -          & - & - & -          & -   & -                                                   & 26.24\%  & 46.17\%         & 36.71\%  & 35.16\%         & \textbf{15.21\%} & 99.32\%   & 97.32\%  \cr

    \bottomrule  
    \end{tabular}  
    \end{threeparttable}  
\end{table*} 

\begin{table}[!t]  
  \centering  
  \fontsize{5}{6}\selectfont  
  \begin{threeparttable}  
  \caption{Quality assessment for anti-forensics DeepFakes generated by different methods} 
  \label{compare_visual}  
    \begin{tabular}{@{\hspace{2pt}}c@{\hspace{2pt}}c@{\hspace{2pt}}c@{\hspace{2pt}}c@{\hspace{2pt}}c@{\hspace{2pt}}c@{\hspace{2pt}}c@{\hspace{2pt}}c@{\hspace{2pt}}c@{\hspace{2pt}}c@{\hspace{2pt}}c@{\hspace{2pt}}c@{\hspace{2pt}}c@{\hspace{2pt}}c@{\hspace{2pt}}c@{\hspace{2pt}}c@{\hspace{2pt}}}  
    \toprule  
    \multirow{2}{*}{Model}&  
    \multicolumn{5}{c}{Celeb-DF} &\multicolumn{5}{c}{DeeperForensics}  &\multicolumn{5}{c}{FF++}\cr  
    \cmidrule(lr){2-6} \cmidrule(lr){7-11} \cmidrule(lr){12-16} 
     &\cite{isola2017image} &\cite{ding2021anti} &\cite{fan2021deepfake} &\cite{wang2021adversarial} &ours
     &\cite{isola2017image} &\cite{ding2021anti} &\cite{fan2021deepfake} &\cite{wang2021adversarial} &ours
     &\cite{isola2017image} &\cite{ding2021anti} &\cite{fan2021deepfake} &\cite{wang2021adversarial} &ours
     \cr
    \midrule
   PSNR                                  &26.81  & 23.16    & 19.12   &30.39 & \textbf{31.49}               &28.49   & 26.58           & 18.08  &\textbf{31.02}  & 28.38          &27.37 & 29.24            & 20.71   &29.65   & \textbf{30.70}  \cr
    SSIM                                  &0.935 & 0.893   & 0.601  &0.928         & \textbf{0.943}         &\textbf{0.942}  & 0.936          & 0.758   &0.931  & 0.929            &0.931 & 0.958 & 0.630  &0.949  & \textbf{0.966}          \cr
    FaceDetection\cite{zhou2013extensive} &0.974  & 0.976    & 0.786   &0.975          & \textbf{0.984}      &0.994   & 0.996        & 0.958   &0.976  & \textbf{0.998}             &0.998 & 0.997            & 0.986   &0.998   & \textbf{0.999}  \cr
    \bottomrule   
    \end{tabular}  
    \end{threeparttable}  
\end{table}

\section{experiment}

\textbf{Datasets:}We extracted frames from videos and located and segmented the facial areas in all images. To ensure consistent sizing, we resampled and cropped the facial images to 256 × 256 pixels for all experiments. Additionally, we manually removed samples with noticeable visual defects from the experimental sets. Our dataset comprised 440,000 pairs of images from Celeb-DF \cite{li2020celeb}, 480,000 pairs from FaceForensics++ \cite{rossler2019faceforensics++}, and 470,000 pairs from DeeperForensics \cite{jiang2020deeperforensics}. Each pair consisted of a real frame and a corresponding DeepFake. 

\textbf{Evaluation Metrics}. We present the classification performance of all models in TABLE \ref{table_detector_accuracy}. All detectors exhibit outstanding classification accuracy, surpassing 97\%. Additionally, we include prediction precision as a supplementary metric, which represents the proportion of correctly predicted fake images.

\subsection{Comparison with other methods}
\textbf{Prior to Assessments:} We utilize the DeepFake images $I_f$ and the DeepFake sharpened images $I_{fu}$ as baselines. For a comprehensive evaluation, we utilize benchmarks such as \cite{isola2017image}, \cite{ding2021anti}, \cite{fan2021deepfake}, and the black-box attacks portion of \cite{wang2021adversarial} for comparison. Once all the models are trained, we utilize them to generate adversarial fake images, which will undergo further testing and evaluation.

\textbf{Comparison of undetectability:} In TABLE \ref{compare_undetected}, we evaluate the performance of adversarial fake images using the pre-trained detectors mentioned in TABLE \ref{table_detector_accuracy}.

It's worth highlighting that the proposed method achieves high levels of undetectability, even on the challenging DeepForensics dataset. DeepForensics introduces greater randomness by incorporating various factors such as random noise, blurring, distortion, and compression to simulate real-world imperfections\cite{jiang2020deeperforensics}. This increased complexity makes achieving high undetectability on DeepForensics particularly valuable.

In comparison to other forensic detectors, SBIs exhibit higher robustness, further demonstrating that self-blended methods can indeed mitigate overfitting to specific subsets of deepfake data\cite{shiohara2022detecting}.  

Notably, all detectors consistently maintain an accuracy rate of over 90\% in detecting $I_{fu}$. This reaffirms the idea discussed in the method's motivation (Fig. \ref{illus_motivation}) that relying solely on traditional USM sharpening is insufficient to effectively deceive the detectors.

\textbf{Evaluation of visual quality:} \textbf{1)} Some samples of generated images are displayed in Fig. \ref{compare_Visual_img}. Many of the samples synthesized by \cite{isola2017image}, \cite{fan2021deepfake}, and \cite{wang2021adversarial} exhibit noticeable visual defects, including color mismatches, distortions, consistent tonal issues, and unreasonable artifacts. As previously emphasized, these defects are unacceptable in the context of anti-forensics, as they are easily detected by human observers. In contrast, most images synthesized by \cite{ding2021anti} and the proposed method are highly convincing, featuring clear facial silhouettes and devoid of any dubious artifacts or unrealistic effects. Moreover, our method enhances image quality through sharpening, rendering it superior to \cite{ding2021anti}.
\textbf{2)} To conduct a comprehensive quality assessment, we employ the Peak Signal-to-Noise Ratio (PSNR) and the Structural Similarity Index (SSIM) as quality metrics in this evaluation. Additionally, we apply the methodology outlined in \cite{zhou2013extensive} to ascertain whether the synthesized faces can be recognized by a face recognition algorithm. The results, presented in TABLE \ref{compare_visual}, demonstrate that the proposed method excels in terms of PSNR, SSIM and face detection, particularly in the Celeb-DF dataset.

\section{Conclusion}
In this paper, we introduce a novel model for black-box anti-forensics of DeepFakes. Our model consists of two sub-networks: the Forensics Disruption Network and the Visual Enhancement Network. Following training, our method generates adversarial sharpening masks to make DeepFake images appear as naturally sharpened ones, achieving both high undetectability and improved image quality.

\bibliographystyle{IEEEbib}
\bibliography{strings,refs}

\end{document}